%% file: main.tex
\documentclass[conference]{IEEEtran}
\IEEEoverridecommandlockouts
\usepackage{cite}
\usepackage{amsmath,amssymb,amsfonts}
\usepackage{algorithmic}
\usepackage{graphicx}
\usepackage{textcomp}
\usepackage{xcolor}
\def\BibTeX{{\rm B\kern-.05em{\sc i\kern-.025em b}\kern-.08em
    T\kern-.1667em\lower.7ex\hbox{E}\kern-.125emX}}

\usepackage{booktabs}
\usepackage{mathtools}
\usepackage{wrapfig}
\usepackage{enumitem}
\usepackage{makecell}
\usepackage{url}

\usepackage{listings}
\usepackage{adjustbox}

\newcommand{\update}[1]{\textcolor{black}{#1}}

\begin{document}

\title{Grammar and Gameplay-aligned RL \\ for Game Description Generation with LLMs
}

\author{Tsunehiko Tanaka, Edgar Simo-Serra
        % <-this % stops a space
\thanks{The authors are with Waseda University, Tokyo, Japan. (Corresponding author: Tsunehiko Tanaka, email: tsunehiko@fuji.waseda.jp)}}

\maketitle

\begin{abstract}
Game Description Generation (GDG) is the task of generating a game description written in a Game Description Language (GDL) from natural language text. Previous studies have explored generation methods leveraging the contextual understanding capabilities of Large Language Models (LLMs); however, accurately reproducing the game features of the game descriptions remains a challenge. In this paper, we propose reinforcement learning-based fine-tuning of LLMs for GDG (RLGDG). Our training method simultaneously improves grammatical correctness and fidelity to game concepts by introducing both grammar rewards and concept rewards. Furthermore, we adopt a two-stage training strategy where Reinforcement Learning (RL) is applied following Supervised Fine-Tuning (SFT). Experimental results demonstrate that our proposed method significantly outperforms baseline methods using SFT alone. Our code is available at https://github.com/tsunehiko/rlgdg
\end{abstract}


\input{introduction}
\input{related_work}
\input{problem_setting}
\input{method}
\input{experiments}
\input{experimental_results}
\input{limitation}
\input{conclusion}

\section*{Acknowledgments}
This work was supported by JST, ACT-X Grant Number JPMJAX23CE, Japan.

\bibliographystyle{IEEEtran}
\bibliography{abrv,reference}

\end{document}

%% file: introduction.tex
\section{Introduction}
Game Description Language (GDL)~\cite{gdl, vgdl, rbg, egd, ludii} is a domain-specific language used to represent a wide variety of games in a unified notation. 
For instance, the Ludii GDL~\cite{ludii} primarily models board games and covers more than 1,000 different game types. 
Game descriptions written in GDL are highly machine-readable, making it easy for dedicated game engines to run simulations. 
Because GDL descriptions can be automatically evaluated through simulation, they have become widely used in automated game design research~\cite{egd, arcade, rule_generation, random_forest}.

Recently, there has been increased interest in Game Description Generation (GDG)~\cite{llmgg, ggdg}. 
GDG focuses on generating game descriptions from natural language texts, making it easier for non-experts to participate in game design. 
In GDG tasks, In-Context Learning (ICL) with Large Language Models (LLMs)~\cite{gpt2} has shown great promise.
LLMs excel in understanding textual contexts and can perform tasks with limited domain knowledge based on a few examples provided in prompts. 
For example, Hu \emph{et al.}~\cite{llmgg} demonstrated that enriching prompts with explanations of GDL notation and examples of game descriptions improves the accuracy of generated outputs. 
Additionally, \update{Grammar-based Game Description Generation (GGDG)}~\cite{ggdg} proposed an iterative decoding method guided by GDL grammar rules, significantly enhancing grammatical correctness. 
However, the iterative improvements in GGDG have restricted grammatical accuracy and do not consider the actual gameplay behavior and features obtained through simulation.

\begin{figure}[t]
    \centering
    \includegraphics[width=0.7\linewidth]{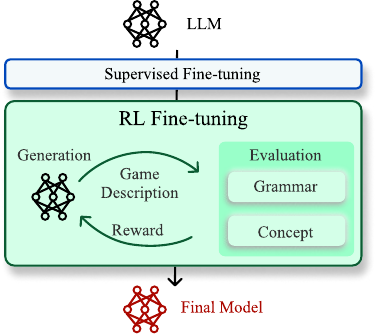}
    \caption{\textbf{Overview of the proposed method.} The training process consists of two stages, starting with Supervised Fine-Tuning followed by Reinforcement Learning-based Fine -Tuning (RLFT), where rewards based on grammar and game concepts are utilized.}
    \label{fig:teaser}
\end{figure}

To address this issue, we propose Reinforcement Learning fine-tuning of LLMs for GDG (RLGDG), aiming to simultaneously enhance grammatical accuracy and fidelity of game features to the ground truth. 
Specifically, we design two types of rewards: (i) a grammatical reward evaluating whether the generated game description complies with GDL grammar, and (ii) a conceptual reward evaluating how accurately the generated game concepts~\cite{board_concept}, such as board cell usage and the proportion of states with multiple possible moves, align with the ground truth. 
As illustrated in Fig.~\ref{fig:teaser}, our proposed method employs a two-stage training procedure: first, Supervised Fine-Tuning (SFT) is conducted, which is then followed by Reinforcement Learning-based Fine-Tuning (RLFT)~\cite{reft, r1, kumar2024training} based on reward optimization. 
By initially learning the basic grammar and structure of game descriptions through SFT, we mitigate unstable outputs in the early stages of RLFT. 
Experimental results demonstrate that our proposed framework outperforms baseline methods not only in grammatical correctness but also in the fidelity of game concepts.

Our main contributions are as follows:
\begin{itemize}
    \item We propose Reinforcement Learning fine-tuning of LLMs for GDG (RLGDG) to jointly enhance grammatical accuracy and game feature fidelity.
    \item Our approach introduces grammatical and conceptual rewards to align generated game descriptions with grammar and actual gameplay features.
    \item We validate through extensive experiments that our proposed method significantly improves GDG performance.
\end{itemize}

%% file: related_work.tex
\section{Related Work}
\subsection{Game Description Language}
\looseness=-1
Game Description Language (GDL) is a specialized language for describing specific games. 
In 2005, GGP-GDL~\cite{gdl} was introduced for General Game Playing, aiming to develop artificial intelligence agents capable of adapting to various games. 
Video Game Description Language (VGDL)~\cite{vgdl} has been developed to represent rules and levels of 2D sprite-based games, currently modeling as many as 195 different games. 
Regular Boardgames (RBG)~\cite{rbg} is another language that combines high-level language features with low-level descriptions, enabling the representation of complex board games. 
The Ludi system~\cite{egd}, incorporating evolutionary game design, successfully led to the development of the commercially successful game ``Yavalath.'' 
Moreover, Ludii~\cite{ludii} is a system adopting the ``ludemic approach,'' allowing the decomposition and description of game components at a conceptual level. 
This enables Ludii to represent over 1,000 traditional games, including board games, card games, dice games, and tile games. 
Given the broad representational capability and versatility of Ludii GDL, we primarily use Ludii GDL for our research.

Game analysis using Ludii is actively progressing, particularly within board game research~\cite{measuring, manual, board_concept, heuristic, universal}. 
For instance, methods~\cite{measuring} have been proposed to quantify similarities between board games using concept values defined in Ludii. 
Moreover, Stephenson \emph{et al.}~\cite{manual} developed a framework utilizing Ludii to automatically generate rule descriptions for board games. 
In this study, we focus specifically on generating game descriptions using Ludii GDL.

\subsection{Large Language Models in Games}
Since the emergence of ChatGPT~\cite{openai_2022}, large language models (LLMs) have garnered significant attention, prompting exploration into diverse applications within the field of game AI~\cite{gallotta2024large, practicalllm}. 
For example, several methods~\cite{levelllm, shyam2023mariogpt} have been proposed for generating 2D tile-based game levels by fine-tuning GPT-2~\cite{gpt2}. 
Additionally, research is progressing on prompt-based LLM methods for level generation~\cite{chatgpt4pcg} and quest generation in role-playing games~\cite{rpg}. 
Dreamcraft~\cite{dreamcraft} further demonstrates a technique to create 3D game objects within Minecraft from textual prompts. 
Li \emph{et al.}~\cite{unbounded} introduces a virtual pet game that achieves real-time gameplay even with smaller LLMs by employing a domain-specific distillation approach. 

\update{On the other hand, research has also advanced in automatically generating game rules and descriptions using LLMs and GDL.}
GAVEL~\cite{gavel} employs an LLM fine-tuned on Ludii game descriptions as a mutation operator in evolutionary search, aiming to create novel games. 
The objective of GAVEL differs from our research, which focuses on generating game descriptions that maintain consistency with natural language text. 
Hu \emph{et al.}~\cite{llmgg} proposes a method for generating both rules and levels in VGDL using LLMs. 
LLMaker~\cite{llmaker} improves content consistency by utilizing function calling, but its creativity is limited to the scope of the defined functions.
GGDG~\cite{ggdg} utilizes iterative decoding based on the grammar of Ludii GDL to enhance the grammatical correctness of generated game descriptions. 
\update{While these previous studies primarily focus on novelty or grammatical correctness of game descriptions, our research differs by emphasizing the improvement of game characteristics of generated game descriptions, bringing them closer to the ground truth through reinforcement learning to enhance LLMs.}

\subsection{RL for LLMs}
\looseness=-1
RL-based fine-tuning of LLMs~\cite{rlhf, r1, reft, kumar2024training} has been attracting attention. 
Reinforcement Learning with Human Feedback (RLHF)~\cite{rlhf} aligns model behaviors with human preferences by using human feedback as rewards. 
Recent methods~\cite{r1, reft} have explored accuracy-based reward functions without human feedback, notably improving logical reasoning tasks such as mathematics and programming. 
RL-finetuned models are known to exhibit enhanced reasoning capabilities~\cite{r1}. 
Additionally, advancements in RL algorithms led to the development of Group Relative Policy Optimization (GRPO)~\cite{grpo}, a variant of Proximal Policy Optimization (PPO)~\cite{ppo} optimized for fine-tuning LLMs. 
In this study, we focus on GDG and define a reward function based on both grammatical correctness and conceptual relevance to the game in generated descriptions, employing GRPO to fine-tune the LLM.

\begin{figure}[t]
    \centering
    \includegraphics[width=\linewidth]{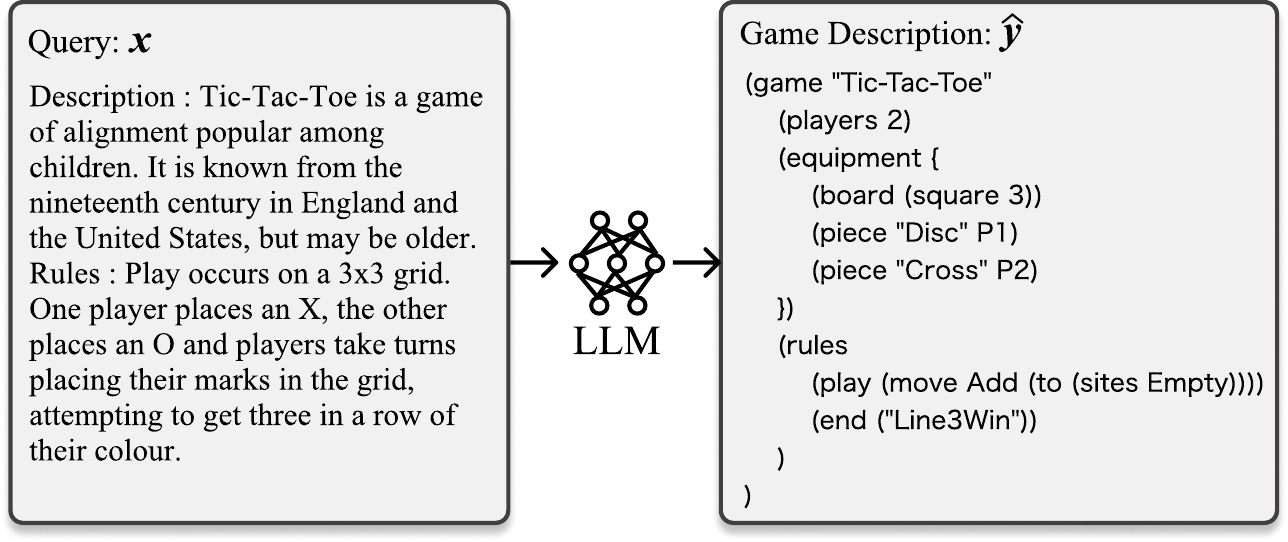}
    \caption{\looseness=-1\textbf{An example of GDG for the game Tic-Tac-Toe.} $x$ is text that explains games in natural language. \update{$\hat{y}$ is the game description generated by the LLM in Ludii GDL, a Game Description Language.}}
    \label{fig:gdg}
\end{figure}

%% file: problem_setting.tex
\section{Problem Setting}
In this section, we define GDG, the primary task addressed in this paper. 
As shown in Fig.~\ref{fig:gdg}, the task involves providing a query $x$ to a large language model (LLM), which then generates the corresponding game description $y$. 
The query $x$ is a natural language sentence describing the content or rules of a game. 
Our goal is to make the game description $\hat{y}$, generated by the LLM, as close as possible to the ground-truth game description $y$. 
\update{The acquisition of query $x$ and ground truth $y$ is described in Section~\ref{subsec:datasets}.}
% In this study, we employ Ludii GDL, a Game Description Language (GDL) commonly used in recent research.
% In this study, we use Ludii GDL as one of the forms of GDL. 
% This is because Ludii GDL is a versatile framework capable of describing a wide variety of classical games, and it has been widely employed in recent research on GDL~\cite{measuring, manual, board_concept, heuristic, universal, ggdg}.

%% file: method.tex
\section{Methodology}
\subsection{Training Procedure}
Our training procedure consists of two steps. 
First, we perform SFT using the paired data of queries and corresponding game descriptions. 
Then, we conduct RLFT starting from the SFT-trained model.

SFT in the first step allows LLMs to avoid unstable outputs commonly encountered in the initial stages of RL. 
For example, to simplify the extraction of generated programs, we instruct the model to output in a specific format, such as \texttt{<program>...</program>}. 
Models without SFT, \update{especially those with smaller parameter sizes such as 1.5B,} often ignore the format instructions, generating extraneous text or incorrect formats like \texttt{\textasciigrave\textasciigrave\textasciigrave xml (program)...\textasciigrave\textasciigrave\textasciigrave}. 
By resolving these formatting issues through the SFT process, the subsequent RL step can focus solely on improving the quality of the game description.

For RLFT, we use GRPO. 
This algorithm is one of the prominent RL methods for LLMs and has been adopted in DeepSeek-R1~\cite{r1}, an LLM renowned for its strong reasoning abilities. 
GRPO extends PPO by using rewards from multiple sampled output candidates $\{o_1, o_2, ..., o_G\}$ for each query, rather than relying on a value function. 
This approach removes the necessity of value function approximation, thereby enhancing training stability. 
We discuss reward modeling methods for GRPO in the following subsection.

\subsection{Reward Modeling for GDG}
\label{subsec:reward}
We design two types of rewards: grammar rewards and concept rewards. 
By employing both, the model can improve not only grammatical accuracy but also the fidelity to the ground truth in the game concept.

% \paragraph{Edit distance rewards}
% Edit distance reward is a metric that uses edit distance to evaluate how close the output $\hat{y}$ is to the correct answer $y$. 
% This reward is responsible for the syntactic-based evaluation. 
% Both $\hat{y}$ and $y$ are tokenized according to the syntactic rules of Ludii GDL, and the token-level edit distance $d$ is computed. 
% Let $T_{\hat{y}}$ be the token length of $\hat{y}$, and $T_y$ be the token length of $y$. 
% The edit distance reward $r_d$ is calculated as follows:
% \begin{align}
%     r_d = 1 - \frac{d}{\max(T_{\hat{y}}, T_y)}.
% \end{align}
% The scale of this reward ranges from 0 to 1, with values closer to 1 indicating greater similarity between $\hat{y}$ and $y$.

\paragraph{Grammar rewards}
\looseness=-1
Grammar rewards measure how much of the output $\hat{y}$ is grammatically valid according to the GDL grammar. 
Using the Earley parser \update{implemented in Ludii~\cite{earley}}, we parse $\hat{y}$ from the beginning based on the GDL grammar and obtain the longest grammatically correct substring $\hat{y}_\mathrm{valid}$.
Let $L_{\hat{y}}$ be the string length of $\hat{y}$ and $L_{\hat{y}_\mathrm{valid}}$ be the string length of $\hat{y}_\mathrm{valid}$. 
The grammar reward $r_g$ is calculated as follows: 
\begin{align}
    r_g = \frac{L_{\hat{y}_\mathrm{valid}}}{L_{\hat{y}}}.
\end{align}
When the length of $\hat{y}_\mathrm{valid}$ equals that of $\hat{y}$, it reaches its maximum value 1. 
This reward scale ranges from 0 to 1, approaching 1 as the grammatically valid portion increases.

\paragraph{Concept rewards}
\begin{figure}[t]
    \centering
    \includegraphics[width=0.6\linewidth]{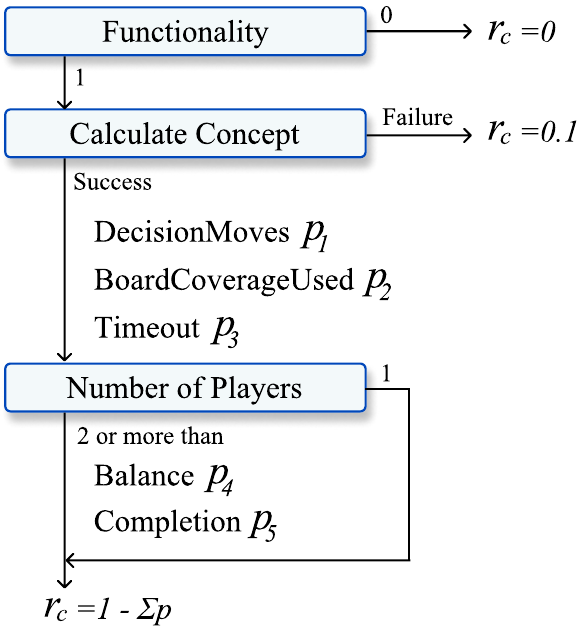}
    \caption{\textbf{Overview of the calculation process for the concept reward.} First, functionality is evaluated, and the concept is calculated only for functional outputs. If the concept can be calculated, penalties $p_i$ are computed across three or five items, depending on the number of players, to derive the final reward $r_c$.}
    \label{fig:concept_reward}
\end{figure}
Concept rewards evaluate the similarity between the game features of the predicted output $\hat{y}$ and those of the ground truth $y$.
An overview of the calculation process for the concept reward is shown in Fig.~\ref{fig:concept_reward}.
% First, we determine whether the output has the necessary playable functionality. 
% If rules fail to function properly—such as when a player cannot move their pieces—the output is considered non-functional, and the reward $r_c$ is set to 0. 
% If the output is functional, we proceed to the next step. 
\update{
First, we determine whether the generated game is functional. 
To clarify this, we introduce two notions: 
Compatibility indicates whether the Ludii game engine can parse and compile the game without errors, and Functionality indicates whether the compiled game works well enough to be played. 
If rules fail to function properly, such as when a player cannot move their pieces, the output is considered non-functional. 
Only compilable games can be evaluated for functionality. 
A functional game proceeds to the next evaluation step; otherwise, the reward $r_c$ is set to 0. }
In the next step, we compute the concept values for the output $\hat{y}$ and compare them with the concept values of $y$.
Concept values represent features of a game and were introduced in \cite{board_concept}. 
Although there are hundreds of concept values, \update{as an initial exploration of incorporating game concepts into RL}, we evaluate the following five items, taking inspiration from GAVEL~\cite{gavel}.
\begin{enumerate}
    \item DecisionMoves $c_1$: Percentage of \update{terns} where there was more than one possible move.
    \item BoardCoverageUsed $c_2$: Percentage of used board sites on which a piece was placed at some point.
    \item Timeout $c_3$: Percentage of games that end via timeout.
    \item Balance $c_4$: Similarity between player win rates.
    \item Completion $c_5$: Percentage of games that have a winner (not a draw or timeout).
\end{enumerate}
Each item is cited from the concept definitions of the Ludii concept search~\cite{ludii_concept}.
Items 1, 2, and 3 are measured for all functional games, while items 4 and 5 are measured only for games with two or more players.
These values are extracted from automatic playouts under a random policy, based on \cite{board_concept, measuring}. 
Following prior studies~\cite{gavel, ggdg}, we perform 50 playouts for the ground truth game and 10 playouts for the predicted game. 
Let the concept values from the output $\hat{y}$ be denoted as $\hat{c}$. 
We compute the penalties for each item using a Gaussian kernel:
\begin{align}
    p_i = 1 - \exp\Bigl(-\frac{1}{2} \bigl(\tfrac{\hat{c}_i - c_i}{\sigma}\bigr)^2\Bigr), 
\quad 1 \leq i \leq 5,
\end{align}
where $\sigma = 0.3$. 
By employing a Gaussian kernel, the scale of each penalty is normalized to a range from 0 to 1. 
The penalty approaches 1 as the features of the output $\hat{y}$ deviate further from those of $y$.
We then use a weighted sum of these penalties to compute the reward:
\begin{align}
    r_c = 1 - \sum_{i=1}^5 w_i \, p_i.
\end{align}
\update{
Here, $w$ denotes the weights. 
When the game is functional but the game features of $\hat{y}$ and $y$ differ greatly—that is, when $c_1, \dots, c_5$ are all equal to $1.0$—we grant a small reward of $r_c = 0.1$ for simply functioning. 
We implement this by setting every $w$ to $0.18$.
Furthermore, if the five concept values cannot be computed—for example, when the game is functional but the automatic playout calculation exceeds the timeout limit—we also set $r_c$ to $0.1$. 
Note that the timeout is set to $180$ seconds.}

Finally, we combine the three rewards as follows:
\begin{align}
    r = r_g + \lambda_c r_c,
\end{align}
where $\lambda_c$ is a scaling parameter, and is set to 1.0.

%% file: experiments.tex
\section{Experiments}
\subsection{Datasets}
\label{subsec:datasets}
We use only game instances from Ludii-1.14~\cite{ludii_portal} whose game description $y$ has a token length of 500 or less for both training and evaluation. 
Token length is calculated using the tokenizer of Qwen2.5-1.5B-Instruct~\cite{qwen25}. 
Our evaluation is performed on 100 randomly selected instances, and the other 410 instances are used for training.

The query $x$ consists of metadata provided by the Ludii game system~\cite{ludii_portal}, specifically description and rules. 
``Description'' gives an overview of the game, while ``Rules'' detail the game's specific rules. 
Following prior work~\cite{gavel, ggdg}, and to enhance the dataset's generality, we use a game description $y$ in which the game-specific functions defined within each game are fully expanded. 
After expansion, the game descriptions rely exclusively on primitive functions, omitting Ludii's meta-language features such as definitions, options, rulesets, ranges, and constants.

\subsection{Comparison Methods}
Here are the methods we compare in our experiments:
\begin{itemize}
    \item \textbf{GDG}: A baseline approach using LLM's ICL without SFT or RL to generate game descriptions. The prompt includes demonstration examples $(x^{(i)}, y^{(i)})_{i=1}^N$, each consisting of a query and corresponding game description. We set $N=3$.
    
    \item \textbf{GGDG}~\cite{ggdg}: A baseline method designed to improve the grammatical correctness of the generated output $\hat{y}$ based on Ludii GDL grammar. It builds upon GDG by iteratively refining $\hat{y}$ through grammar-based decoding.
    
    \item \textbf{SFT+GDG}: A baseline approach where the LLM is fine-tuned using SFT alone. The model is trained on the dataset described in Section~\ref{subsec:datasets}.
    
    \item \textbf{RLGDG (ours)}: Our proposed approach, which employs an LLM first fine-tuned with SFT and then further fine-tuned with RL. In RLFT, we use the same dataset as the SFT step, as described in Section~\ref{subsec:datasets}.
\end{itemize}
The GDG and GGDG methods perform few-shot inference using demonstration examples, whereas SFT+GDG and RLGDG perform zero-shot inference without demonstration examples.

\subsection{Implementation Details}
\begin{figure}[t]
    \centering
    \includegraphics[width=0.9\linewidth]{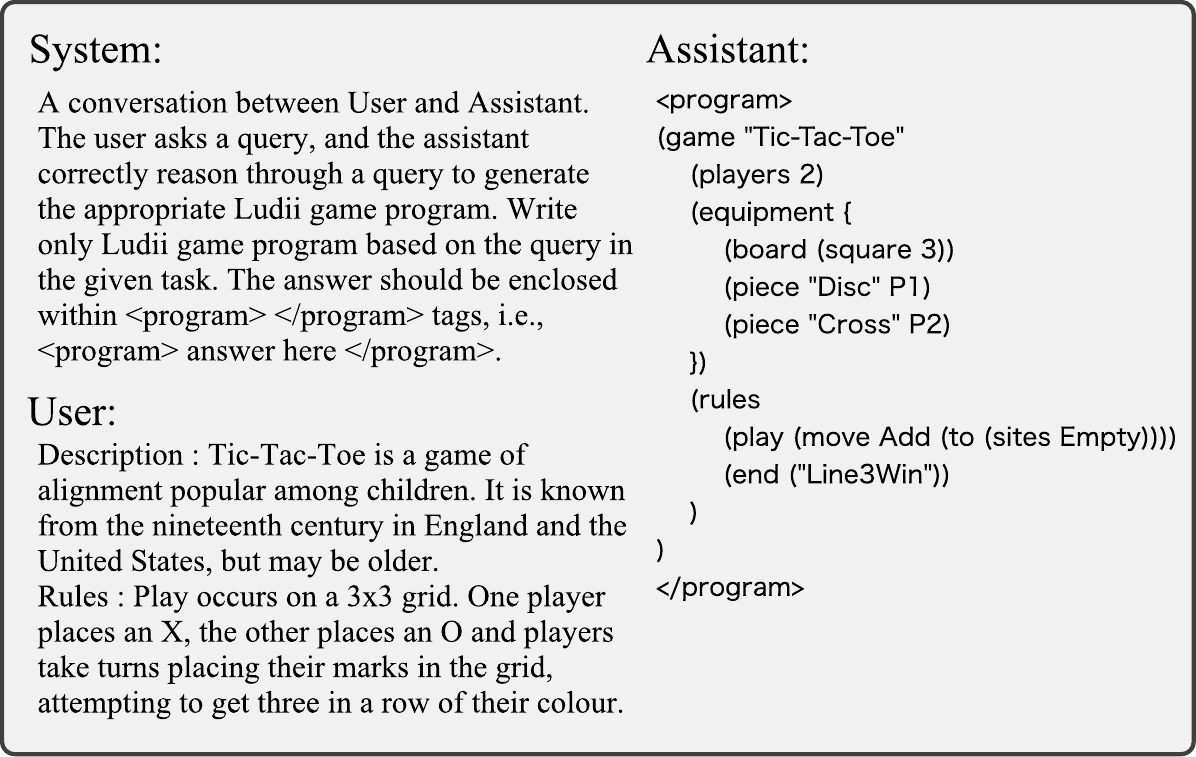}
    \caption{\textbf{Our prompts for SFT and RLFT.}}
    \label{fig:prompt}
\end{figure}
\looseness=-1
We utilize Qwen2.5-1.5B-Instruct~\cite{qwen25} as the LLM. 
This choice is motivated by two primary reasons: first, we select an open-source LLM to ensure reproducibility of the research results. 
Second, the chosen model size enables practical fine-tuning with available computational resources. 
Our experiments are carried out using two NVIDIA RTX 6000 Ada GPUs.

\looseness=-1
Both SFT and GRPO methods employ full-parameter fine-tuning. 
Specifically, for the SFT method, we set the sequence length to 768 tokens, batch size to 2, learning rate to 1e-4, and warmup ratio to 0.03, conducting training for a total of 3 epochs.

\looseness=-1
For GRPO, the prompt length is fixed at 256 tokens, and the completion length at 512 tokens. 
We use a batch size of 1, a learning rate of 3e-6, and a warmup ratio of 0.1. 
Furthermore, the number of generated outputs per query $G$ is set to 4. 
We conduct training for a single epoch, using a temperature of 0.9 for sampling output candidates.
Our prompts are shown in Fig~\ref{fig:prompt}.

\looseness=-1
The Earley Parser is implemented using the Lark library~\cite{lark_parser}. 
The grammar for parsing Ludii descriptions strictly adheres to the specifications provided in the Ludii Language Reference.

%% file: experimental_results.tex
\section{Experimental Results}
\subsection{Evaluation Metrics}
\update{To evaluate the generated game descriptions, we employ the following metrics based on prior work~\cite{ggdg}. 
For details on Compilability and Functionality, see Section~\ref{subsec:reward}.}
\begin{itemize}
    \looseness=-1
    \item \textbf{Compilability}: \update{The proportion of games that can be successfully parsed and compiled by the Ludii game engine. This score is normalized to a range from 0 to 100.}
    \item \textbf{Functionality}: \update{The proportion of games considered playable. This score is normalized to a range from 0 to 100.}
    \item \textbf{ROUGE-L}~\cite{rouge}: A metric used to measure linguistic similarity between the generated outputs and ground truth. Commonly used in program synthesis, higher values indicate greater similarity. This metric does not account for grammatical correctness or game-specific features but solely relies on textual similarity. We report the average ROUGE-L F1 score calculated across all test data.
    \item \textbf{Normalized Concept Distance (NCD)}: A metric to measure the similarity of game features between generated outputs and ground truth. Based on \cite{board_concept, measuring}, games are represented as concept value vectors, and their cosine distance is calculated to determine NCD. The concept value vector includes semantic features and behavior data from random playouts, such as the proportion of board positions used at least once, or the proportion of turns where at least one legal move exists. It also includes the five concept rewards described in Section~\ref{subsec:reward}. For ground truth games, 50 playouts are executed, while for generated games, 10 playouts are performed. For non-functional games where the concept distance cannot be calculated, NCD is set to 1. The average NCD is computed to evaluate the quality of GDG.
\end{itemize}

Experiments are conducted using three different random seeds \update{following \cite{ggdg}}, and the mean and standard error for each metric are reported.

\subsection{Comparison with Baseline Methods}
\begin{table}[t]
    \setlength{\tabcolsep}{1mm}
    \centering
    \caption{\textbf{Comparison with baseline methods.} The best results are in \textbf{bold}.}
    \begin{tabular}{lrrrr}
    \toprule
    Method & Compilability$\uparrow$ & Functionality$\uparrow$ & ROUGE-L$\uparrow$ & NCD$\downarrow$ \\
    \midrule
    GDG & 24.3$\pm$3.0 & 23.3$\pm$2.4 & 53.0$\pm$0.6 & 0.74$\pm$0.04 \\
    GGDG & 11.7$\pm$0.7 & 10.3$\pm$0.7 & 29.5$\pm$3.1 & 0.81$\pm$0.02 \\
    SFT+GDG & 54.3$\pm$1.2& 52.0$\pm$1.5 & 60.9$\pm$0.2 & 0.51$\pm$0.01 \\
    RLGDG (ours) & \textbf{71.3$\pm$0.9} & \textbf{70.3$\pm$1.3} & \textbf{64.0$\pm$0.2} & \textbf{0.33$\pm$0.01} \\
    \bottomrule
  \end{tabular}
  \label{tab:main_results}
\end{table}
Table~\ref{tab:main_results} shows the comparative results with baseline methods. 
RLGDG outperforms baseline methods across all evaluation metrics. Notable improvements are observed in Compilability and Functionality, demonstrating that RLGDG effectively ensures grammatical correctness and practical playability of the generated game descriptions. 
Additionally, a significant improvement in NCD suggests that RLGDG enables LLMs to better learn game concepts.

GGDG underperforms compared to GDG across all metrics. 
This is likely because GGDG's iterative improvement decoding requires handling numerous detailed instructions, a task challenging for small-scale models with approximately 1.5B parameters. 
In contrast, RLGDG significantly enhances performance using the 1.5B-parameter model, suggesting that it also holds advantages in terms of inference cost efficiency.

\subsection{Ablation Study}
\noindent \textbf{Comparison of Reward Modeling.}
\begin{table*}[t]
    \centering
    \caption{\textbf{Comparison of reward modeling in our proposed method.} The best results are in \textbf{bold}.}
    \begin{tabular}{lccrrrr}
    \toprule
    Method & Grammar & Concept & Compilability$\uparrow$ & Functionality$\uparrow$ & ROUGE-L$\uparrow$ & \makecell{Normalized \\ Concept Distance$\downarrow$} \\
    \midrule
    SFT+GDG~\cite{ggdg} & & & 54.3$\pm$1.2& 52.0$\pm$1.5 & 60.9$\pm$0.2 & 0.51$\pm$0.01 \\
    RLGDG w/o concept & $\checkmark$ & & 69.0$\pm$1.7 & 66.3$\pm$1.8 & \textbf{64.0$\pm$0.1} & 0.37$\pm$0.02 \\
    RLGDG & $\checkmark$ & $\checkmark$ & \textbf{71.3$\pm$0.9} & \textbf{70.3$\pm$1.3} & \textbf{64.0$\pm$0.2} & \textbf{0.33$\pm$0.01} \\
    \bottomrule
  \end{tabular}
  \label{tab:reward_ablation}
\end{table*}
We conduct an ablation study on the reward modeling of RLGDG, and the results are presented in Tab.~\ref{tab:reward_ablation}. 
The results demonstrate that using both types of rewards simultaneously yields the best performance across all evaluation metrics.

Focusing on compilability and functionality, we observe that grammar reward accounts for a significant proportion of the performance improvement from SFT+GDG to RLGDG, specifically 86.5\% for compilability and 87.7\% for functionality. 
This suggests that the grammar reward  contributes to enhancing grammatical accuracy.

Introducing the concept reward results in an additional improvement of 10.8\% in NCD compared to RLGDG without the concept reward. 
This indicates that the inclusion of the concept reward not only helps generate compilable and functional code but also improves the generation of game descriptions that more precisely capture the ground truth game concepts.

\noindent \textbf{Game Category of Test Games.}
\label{subsubsec:category}
\begin{table}[t]
    \setlength{\tabcolsep}{1mm}
    \centering
    \caption{\textbf{Comparison of test instance categories.}}
    \begin{tabular}{lrrrrr}
    \toprule
    Method & Compilability$\uparrow$ & Functionality$\uparrow$ & ROUGE-L$\uparrow$ & NCD$\downarrow$ \\
    \midrule
    \multicolumn{3}{l}{\footnotesize{\textit{board/race}}} \\
    \addlinespace[3pt]
    SFT+GDG & 63.0$\pm$9.8 & 59.3$\pm$9.8 & 51.8$\pm$2.0 & 0.45$\pm$0.09 \\
    RLGDG & 81.5$\pm$3.7 & 74.1$\pm$7.4 & 53.4$\pm$0.7 & 0.29$\pm$0.06 \\
    \midrule
    \multicolumn{3}{l}{\footnotesize{\textit{board/sow}}} \\
    \addlinespace[3pt]
    SFT+GDG & 36.1$\pm$7.3 & 36.1$\pm$7.3 & 71.1$\pm$1.2 & 0.66$\pm$0.07 \\
    RLGDG & 58.3$\pm$16.7 & 58.3$\pm$16.7 & 71.7$\pm$1.6 & 0.45$\pm$0.15 \\
    \midrule
    \multicolumn{3}{l}{\footnotesize{\textit{puzzle}}} \\
    \addlinespace[3pt]
    SFT+GDG & 16.7$\pm$4.8 & 11.1$\pm$7.3 & 40.6$\pm$1.5 & 0.90$\pm$0.07 \\
    RLGDG & 27.8$\pm$10.0 & 22.2$\pm$11.1 & 43.0$\pm$2.3 & 0.80$\pm$0.10 \\
    \midrule
    \multicolumn{3}{l}{\footnotesize{\textit{board/space/line}}} \\
    \addlinespace[3pt]
    SFT+GDG & 76.2$\pm$3.1 & 75.0$\pm$2.1 & 73.9$\pm$0.5 & 0.27$\pm$0.02 \\
    RLGDG & 89.3$\pm$5.5 & 88.1$\pm$6.6 & 76.7$\pm$0.6 & 0.14$\pm$0.06 \\
    \midrule
    \multicolumn{3}{l}{\footnotesize{\textit{board/war}}} \\
    \addlinespace[3pt]
    SFT+GDG & 66.7$\pm$0.0 & 66.7$\pm$0.0 & 58.8$\pm$0.3 & 0.38$\pm$0.00 \\
    RLGDG & 88.9$\pm$0.0 & 83.3$\pm$3.2 & 64.8$\pm$0.6 & 0.23$\pm$0.03 \\
    \bottomrule
  \end{tabular}
  \label{tab:test_category_results}
\end{table}
We investigate the performance comparison across different game categories for test instances. 
Following the methodology of previous research (GGDG), we compared five categories: racing games (board/race), mancala games (board/sow), puzzle games (puzzle), line games (board/space/line), and war games, including capture games (board/war). 
The test instances used here are extracted from the instances used in Section 5.b. The results are summarized in Tab.~\ref{tab:test_category_results}. 
RLGDG outperformes SFT+GDG in all metrics across all categories.

\begin{table}[t]
    \centering
    \caption{\textbf{Number of training instances in each category and the average concept distance from the largest category}.}
    \begin{tabular}{lcc}
    \toprule
    Category & Number of Instances & \makecell{Distance from \\board/space/line} \\
    \midrule
    board/race & 13 & 0.068 \\
    board/sow & 33 & 0.081 \\
    puzzle & 13 & 0.098 \\
    board/space/line & 113 & 0.000 \\
    board/war & 65 & 0.059 \\
    \bottomrule
  \end{tabular}
  \label{tab:train_stats}
\end{table}

For comparing categories, SFT+GDG demonstrates the lowest performance in the puzzle category and the highest performance in the board/space/line category across all metrics. 
We believe this performance difference arises from the varying amounts of training data. 
Table~\ref{tab:train_stats} summarizes the number of training instances in each category, and the average concept distance from the board/space/line category, which has the most training instances. 
As the board/space/line category has the largest number of instances, it is considered easier for the model to learn from. 
In contrast, the puzzle category has the fewest instances, equal in number to the board/race category. 
When compared to board/race games, puzzle games are conceptually farther from board/space/line than board/race games. 
Therefore, it is easier for the model to transfer insights gained from the board/space/line category to the board/race category than to the puzzle category, leading to lower performance in the puzzle category.

For comparing methods, RLGDG improves compilability by 66.5\%, functionality by 100\%, and other metrics compared to SFT+GDG in the puzzle category. 
Furthermore, in the board/space/line category, where SFT+GDG already demonstrated high performance, RLGDG further improved performance, achieving an NCD of 0.14. 
These results suggest that RLGDG may overcome the limitations of SFT independently of the category or the amount of training data.

\section{Qualitative Analysis}
\begin{figure*}[t]
    \centering
    \includegraphics[width=0.925\linewidth]{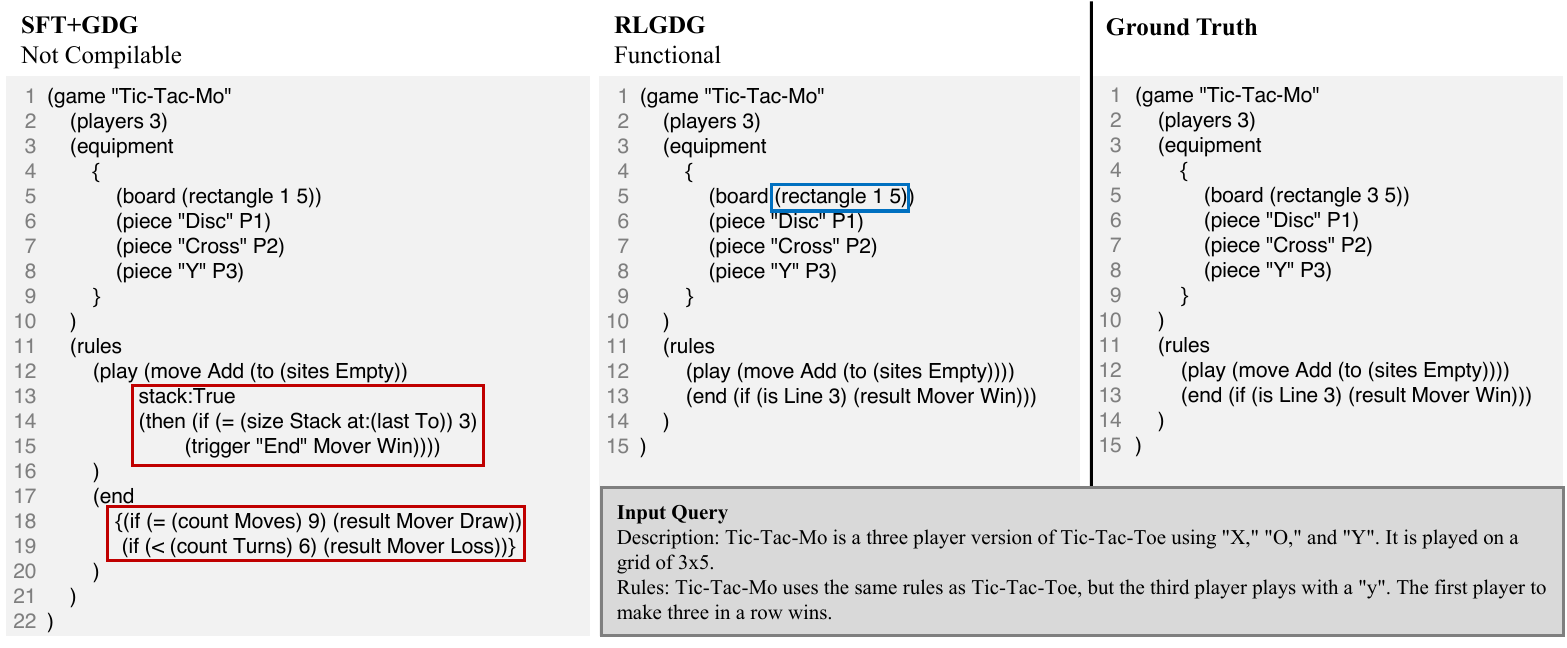}
    \caption{\textbf{Comparison of generation results with baseline methods for Tic-Tac-Mo.} The part enclosed in red differs from the ground truth.}
    \label{fig:tic_tac_mo}
\end{figure*}

\noindent \textbf{Comparison with Baseline Methods.}
We conduct a qualitative analysis comparing the best-performing baseline method, SFT+GDG, against our proposed method, RLGDG. 
Figure~\ref{fig:tic_tac_mo} shows the generation results for Tic-Tac-Mo, a game that extends the player count of Tic-Tac-Toe to three players.
The result from SFT+GDG is non-compilable and non-functional due to the \texttt{trigger ``End'' Mover Win}. 
According to the grammar rules of Ludii GDL, \texttt{trigger ``End'' Mover} should only be followed by a \texttt{then} clause. 
Additionally, the description regarding the termination conditions, marked with a red rectangle, should be placed within \texttt{end} clause of the rules.
In contrast, the result obtained with RLGDG is compilable and functional, and the output game description closely matches the ground truth. 
The only deviation from the ground truth is the board specification, which is \texttt{rectangle 1 5}, identical to the one from SFT+GDG. 
Using different seeds, it can sometimes result in \texttt{rectangle 3 5}, in which case it perfectly matches the ground truth.

\begin{figure*}[t]
    \centering
    \includegraphics[width=0.925\linewidth]{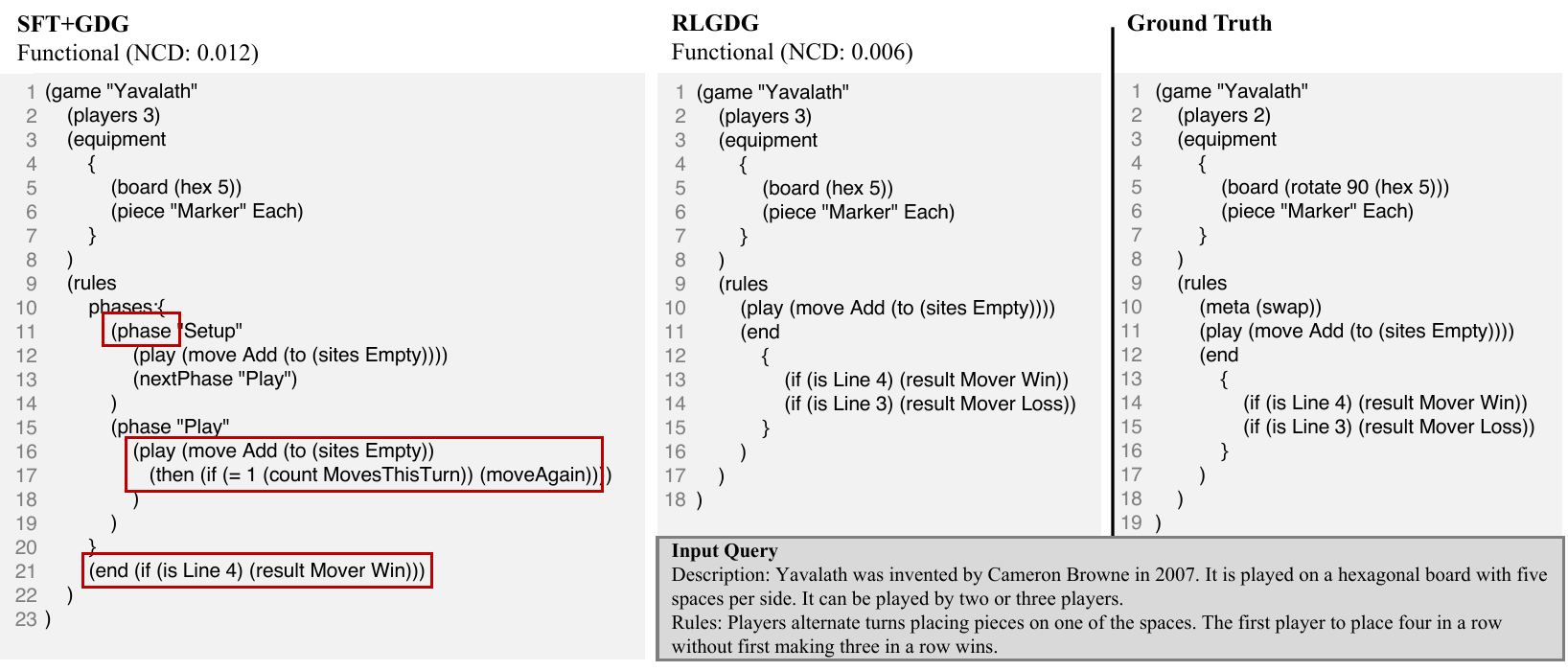}
    \caption{\textbf{Comparison of generation results with baseline methods for Yavalath.} The incorrectly predicted parts are enclosed in red.}
    \label{fig:Yavalath}
\end{figure*}
Figure~\ref{fig:Yavalath} shows the generated results for Yavalath. 
Yavalath is a game developed by the Ludi system~\cite{egd}, where the objective is to align four markers of the same player in a straight line without first aligning three markers. 
The results from SFT+GDG are compilable and functional, with an NCD of 0.012. However, there are primarily three incorrect predictions: 
(i) Introduction of phases not described in Yavalath's rules, 
(ii) inclusion of a move called \texttt{moveAgain}, which allows placing a second marker within the same turn during the \texttt{Play} phase, although no such rule exists in Yavalath, 
and (iii) omission of the termination condition where aligning exactly three markers in a straight line results in a loss.
In contrast, the results from RLGDG are compilable and functional, achieving an NCD of 0.006, indicating that the resulting game concept is closer to the ground truth compared to the SFT+GDG results. 
While the ground truth includes rules such as \texttt{(rotate 90 ...)} and \texttt{(meta (swap))}, these instructions were not included in the input text. 
Additionally, the number of players is different, but both two and three-player options are mentioned in the input text. 
Thus, the game rules described in the text are sufficiently covered by the RLGDG output.

\noindent \textbf{Analysis of Failure Cases.}
\begin{figure*}[t]
    \centering
    \includegraphics[width=0.925\linewidth]{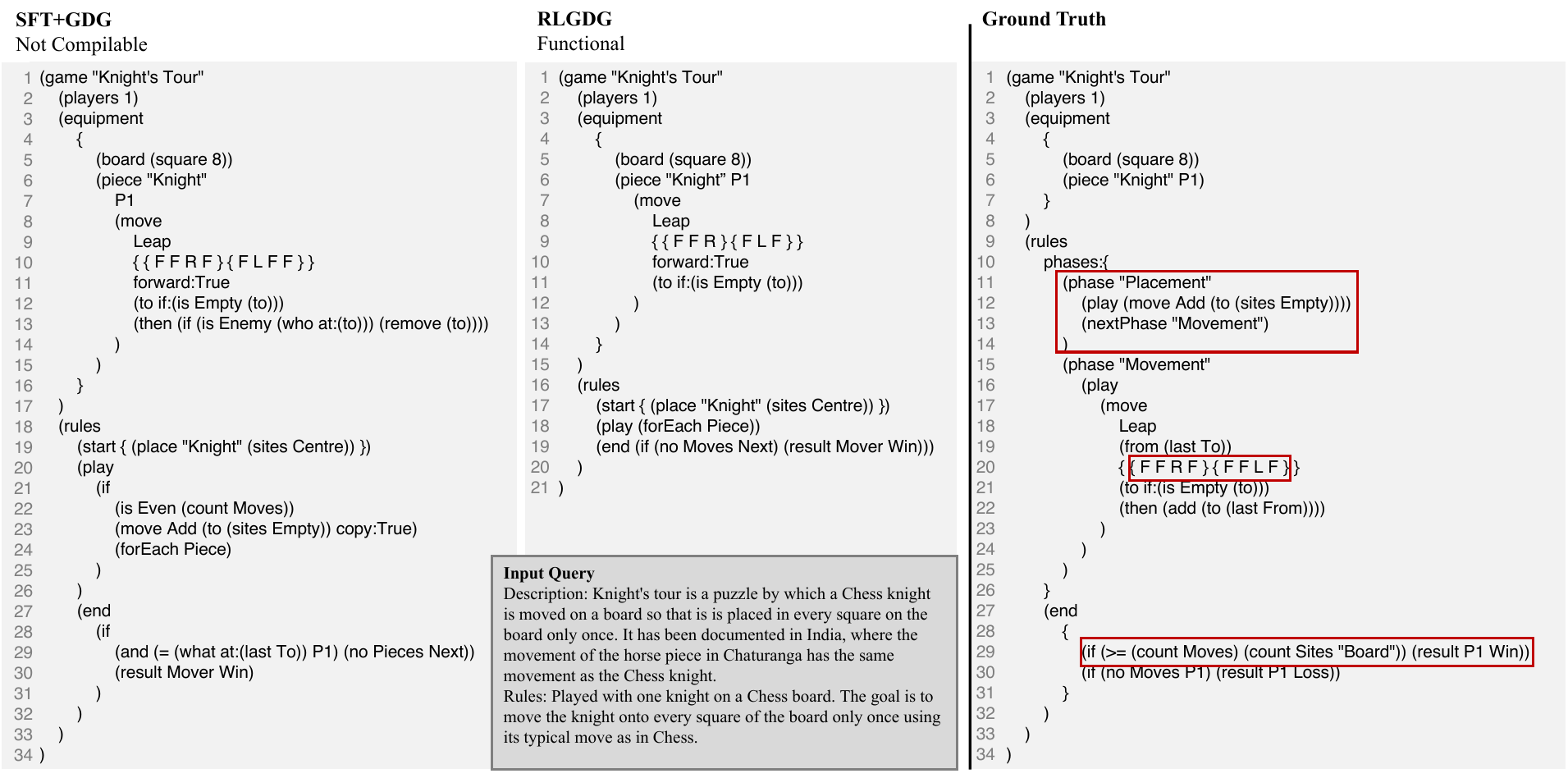}
    \caption{\textbf{Comparison of generation results for Knight's Tour.} Parts of the ground truth that are missing from or incorrectly predicted in the RLGDG output are indicated by boxes.}
    \label{fig:knight's_tour}
\end{figure*}
As an example of a failure case, we analyze the generation results for the Knight's Tour puzzle from the puzzle category, as illustrated in Fig.~\ref{fig:knight's_tour}. 
In this game, the knight is moved across a chessboard so that it visits every square exactly once. 

SFT+GDG generates an output that could not be compiled. 
This was because after the expression \texttt{(if (is Even (count Moves))}, a function returning a boolean value was expected. 
However, it was actually followed by the in-game action \texttt{(move Add (to (sites Empty)) copy:True)}.

The output from RLGDG is compilable and functional but contains errors in the following three aspects: 
(i) In the ground truth, the knight's initial position can be any available square during the placement phase. 
However, the output from RLGDG restricts the knight's initial placement to the central square of the board. 
(ii) The RLGDG output lacks the termination condition at the end of the game, where a win occurs if the knight successfully visits all squares exactly once. 
(iii) The knight's movement pattern generated by RLGDG is incorrect. 
\texttt{F} indicates moving forward by one step, and \texttt{R} means rotating 90 degrees to the right.
In the RLGDG output, the sequence \texttt{FFR} turns the knight right after moving forward two steps. 
However, the correct movement for the knight would require an additional forward step, making the correct sequence \texttt{FFRF}. 
This error likely arises from the limited availability of training data related to the puzzle category, as discussed in Section~\ref{subsubsec:category}.

%% file: limitation.tex
% \section{Limitations and Discussion}
% As illustrated in Section~\ref{subsubsec:category} and Fig.~\ref{fig:knight's_tour}, improvements provided by RLGDG are limited in categories with smaller amounts of training data. 
% Introducing training methodologies utilizing synthetic data~\cite{self-instruct} could potentially offer effective improvements even in categories with limited training data. 
% Additionally, employing evolutionary approaches~\cite{gavel} to generate entirely new game data also represents a promising strategy for augmenting data volume.
% In the experiments, we utilized a relatively small-scale LLM with a parameter size of 1.5B. 
% It is possible that LLMs with larger parameter sizes, such as 70B, could achieve better performance. 
% However, it should be noted that RLFT applied to LLMs with larger parameters demands significantly greater computational resources.

%% file: conclusion.tex
\section{Discussion and Conclusion}
\looseness=-1
In this study, we propose Reinforcement Learning-based fine-tuning of LLMs for Game Description Generation (RLGDG). 
Existing approaches have primarily focused on improving grammatical accuracy; however, our method simultaneously enhances both grammatical correctness and conceptual fidelity to game concepts. 
Specifically, we introduce grammar and concept rewards and adopt a two-stage training strategy that applies RL after Supervised Fine-Tuning (SFT).
Experimental results demonstrated that our proposed method achieved superior performance compared to baseline methods in terms of both grammatical accuracy and conceptual fidelity. 
However, improvements are limited in categories with insufficient training data. 
Future research directions for training data include synthetic data generation~\cite{self-instruct}, data augmentation using evolutionary algorithms~\cite{gavel}, and leveraging larger-scale language models. 
\update{Moreover, extending the concept reward from five to the full set of concept values remains a key target.}
These studies are expected to generate high-quality game descriptions from natural language, thereby supporting designers and engineers in AI-driven game development.